# A game method for improving the interpretability of convolution neural network

**Jinwei Zhao[ab], Qizhou Wang[ab], Fuqiang Zhang[c], Wanli Qiu[c], Yufei Wang[ab], Yu Liu[ab], Guo Xie[ab], Weigang Ma[ab], Bin Wang[ab], Xinhong Hei[ab+]**

[a]Faculty of Computer Science and Engineering, Xi'an University of Technology, Xi'an 710048, China
[b]Shaanxi Key Laboratory of Network Computing and Security Technology (Xi'an University of Technology), Xi'an 710048, China
[c]Faculty of Science, Xi'an University of Technology, Xi'an 710048, China
+ Corresponding author: Xinhong Hei
E-mail: xinghonghei@xaut.edu.cn

**Abstract**

Real artificial intelligence always has been focused on by many machine learning researchers, especially in the area of deep learning. Although deep neural network has surpassed human performance in some specific tasks, it does not have some characters of human intelligence. It is hard to be understood and explained, and sometimes, even metaphysics. It leads to that it is very difficult to remedy its structure and believe its behavior. The reason is, we believe that: the network is essentially a perceptual model. It's like the right hemisphere of human brain is only responsible for dealing with appearance information and carrying out concrete imaginal thinking. Meanwhile, the left hemisphere of the brain is the center of carrying on abstract logic thinking. It has the ability to extract interpretable knowledge from the perception information in the right hemisphere. Therefore, we believe that in order to complete complex intelligent activities from simple perception, it is necessary to construct another interpretable logical network to form accurate and reasonable responses and explanations to external things. Researchers like Bolei Zhou and Quanshi Zhang have found many explanatory rules for deep feature extraction aimed at the feature extraction stage of convolution neural network. However, although researchers like Marco Gori have also made great efforts to improve the interpretability of the fully connected layers of the network, the problem is also very difficult. This paper firstly analyzes its reason. Then a method of constructing logical network based on the fully connected layers and extracting logical relation between input and output of the layers is proposed. The game process between perceptual learning and logical abstract cognitive learning is implemented to improve the interpretable performance of deep learning process and deep learning model. The benefits of our approach are illustrated on benchmark data sets and in real-world experiments.

## Introduction

Real artificial intelligence always has been focused on by many machine learning researchers, especially in the area of deep learning. Although the deep neural network has surpassed human performance in some specific tasks such as AlphaGo and ImageNet, it does not have some characters of human intelligence, like active thinking, independent learning, and associative memory(Wang 2018), and it is hard to be understood and explained, and sometimes, even metaphysics(Rahimi et al. 2017, Brandon et al. 2017). However, in many industries and scenes, for instance in medical and healthcare use cases, it is imperative to interpret decision-makings of deep learning to patients and their families.

The reason is, we believe that: deep learning network is essentially a perceptual network. It's like the right hemisphere of human brain is only responsible for dealing with appearance information such as images and audio information, and carrying out concrete imaginal thinking and divergent thinking. Meanwhile, the left hemisphere of the brain is the center of processing language, carrying on abstract logic thinking, concentrating thinking and analyzing thinking, and has the functions of continuity, order and analysis. It has the ability to extract interpretable knowledge from the perception information of the right hemisphere. Therefore, we believe that in order to complete complex intelligent activities from simple perception, it is necessary to construct another interpretable logical network to form accurate and reasonable responses and explanations to external things, so as to thoroughly solve the interpretable problems in the process of deep learning.

Bolei Zhou et al. (2014, 2016, 2019), Quanshi Zhang et al. (2018, 2019), Runjin Chen et al. (2019) and other researchers have found many explanatory rules for deep feature extraction in their work aimed at the feature extraction stage of a convolution neural network (CNN). However, the interpretation of fully connected layers of the network is always a difficult problem. Many machine learning researchers such as Professor Marco Gori et al. have also made great efforts to improve the interpretability of the fully connected layers(Giannini et al. 2019, Marra et al. 2019a, Marra et al. 2019b), but this problem is still not completely solved.

This paper analyzes the reason why fully connected layers of convolution neural network are unexplainable. Then a method of constructing logical network based on fully connected layers and extracting logical relation between input and output of fully connected layers, Deep Cognitive

Learning Model (DCLM), is proposed. A game process between perceptual learning and logical abstract cognitive learning is implemented to improve the interpretable performance of deep learning process and deep learning model.

## Related works

Building a bridge between logic and learning is a key to constructing a flexible and interpretable perceptual model (Diligenti et al. 2012). Many perceptual learning techniques based on statistical relational learning theory (SRL) have been proposed to act as a bridge between logic and learning. The kFoil algorithm(Landwehr et al. 2006) implemented a dynamic propositionalization approach with kernel methods. Melacci et al.(2013) proposed a box kernel that incorporates supervised points and supervised sets. Veillard et al. (2011) proposed a method for the incorporation of prior knowledge via an adaptation of the standard radical basis function(RBF) kernel. All the above methods solve regression and classification problems by applying relation rules to a kernel method.

Some scientists directly incorporated relation rules into the learning mechanism (Laurer et al. 2009). Fung et al. (2002) proposed incorporating prior knowledge into a linear perceptual classifier in the form of convex constraints in the input space. Maclin et al.(2007) refined the incorrect knowledge and incorporated correct prior knowledge into an perceptual classifier. Diligenti et al.(2010) proposed a general framework to convert first-order logic (FOL) clauses to constraints on real-valued functions by T-norms and incorporate the constraints into a semi-supervised multitask learning scheme. Gori et al.(2011)s introduced equivalent constraints, a constraint checking problem, support constraints, and a constraint induction mechanism in semi-supervised learning problems. Gori et al. (2013) also proposed a general scheme for constraint verification using kernel machines and applied the framework of learning to infer new constraints from old constraints based on kernel-based representations. Maggini et al. (2012) introduced a selection criterion based on a Gauss function in the penalty function to reduce the constraint effect on some points that yield an exception. Zhao et al. (2018) embedded incomplete fuzzy relationships between attributes to a perception learning process. Marra et al. (2019) proposed presents Deep Logic Models which can integrate deep learning and logic reasoning both for learning and inference. All the above methods usually require certain prior knowledge rather than discovering knowledge from the perceptron.

Fan (2018) proposed a generalized hamming network to re-interpret many useful neural network techniques in terms of fuzzy logic. Zhang et al. (2019) proposed a method to learn a decision tree to quantitatively explain the logic of each prediction of a pretrained CNN and mines potential decision modes memorized in fully-connected layers. These methods can find some interpretable knowledge from the CNN. However, they do not implement game process between deep network learning and logic network learning. Then these methods difficultly form accurate and reasonable responses and explanations to external things.

## Theoretical Basis

Suppose X is a compact domain or a manifold in Euclidean space and $Y \in R^k$, $\rho$ is a Borel probability measure of a space $Z = X \times Y$. $f_\rho: X \to Y$ as $f_\rho(x) = \int_Y y d\rho(y|x)$ is defined. The function $f_\rho$ is a regression function of $\rho$. In machine learning, $\rho$ and $f_\rho$ are unknown. At some conditions, an edge probability measure $\rho_X$ of $X$ is known. The number of sample data is m.

The goal of the learning is to find the best approximation of $f_\rho$ in a square integrable function space $\mathcal{H}$ based on a convolution neural network, and extract a logical relation between input and output of the fully connected layers which can interpret the best approximation. Meanwhile, the logical relation can constrain the optimal process of the convolution neural network for following the relation. In this game, the optimal deep neural network is more in line with the people's cognitive laws and the final logical relation can better expresses the essence law of the optimal approximation. Thus these processes will further enhance interpretability.

### Deep Perception Network

In the paper, the predictive process of standard convolution neural network can be decomposed into two perception stages: a feature extraction stage and decision stage. The feature extraction stage includes all the processes before all final fully connected layers as while as the decision stage mainly refers to all final fully connected layers.

In the feature extraction stage, many convolution layers, pooling layers and dropout layers are carried out. Some final feature maps, $\tau_1, \tau_2, \cdots, \tau_k$, are captured from the input vector x from the external world and k is the number of them. We assume that all these feature maps come from a feature space $\Gamma$.

In the decision stage, the feature maps are inputted into all final fully connected layers. The final output, a vector y, is obtained which indicates a classification of input vector x of the convolution neural network.

The Bayes rule allows linking the probability of the parameters to the posterior and prior distributions:

$$p(w|Y_t, X) = \frac{p(Y_t|w,X)p(w|X)}{p(Y_t|X)} \quad (1)$$

We can assume that w which is a parameter vector of the network and $X$ are independent. Then, $p(w) = p(w|X)$. And also because $p(Y_t|X) = 1$, we can obtain

$$p(w|Y_t, X) = p(Y_t|w, X)p(w) \qquad (2)$$

$y_t$ is a target output of the neural architectures. The condition probability $p(y_t|w, x)$ obeys normal distribution

$$p(y_t|w, x) = \frac{1}{Z(y_t)} exp\left(-\frac{(f(w,x)-y_t)^2}{2\sigma^2}\right) \qquad (3)$$

where $f(w, x)$ is the output of the network.
If there are m training samples,

$$p(Y_t|w, X) = \prod_i^m p(y_{ti}|w, x_i) \qquad (4)$$

Training can be carried out by maximizing the likelihood of the training data:

$argmax_w \log p(w|Y_t, X) =$

$argmax_w \left[-\frac{1}{m}\sum_{i=1}^{m}(f(w, x_i) - y_{ti})^2 + \log p(w)\right] \qquad (5)$

where $Z(y_t)$ is assumed as a constant.
Assuming that the parameter vector $w$ priors follow standard Gaussian distributions, we get

$argmax_w \log p(w|Y_t, X) = argmax_w \left[-\frac{1}{m}\sum_{i=1}^{m}(f(w, x_i) - y_{ti})^2 - \frac{\alpha}{2}\|w\|^2\right] \qquad (6)$

If a lot of well training samples can be supplied, an optimal convolution neural network can be obtained by a training process. The optimal convolution neural network can extract the optimal feature maps and predict the correct classification result for all input samples. However, the usual learning network is so complicated that the network cannot be understood by human. It lead to that it is very difficult to remedy its network structure and believe its behavior. Especially, how to improve the interpretation of the fully connected layers of the network is always a difficult problem although many machine learning researchers such as Professor Marco Gori et al. has also made great efforts for solving it (Giannini et al. 2019, Marra et al. 2019a, Marra et al. 2019b).

The reasons for the uninterpretable of the fully connected layers of the convolution neural network are mainly in the following aspects in addition to the complex structure of the fully connected layers. Firstly, in practice, occurrence of random events always causes deviation of measurement data, which generates many intermittent or continuous noise data, so as to make the final neural network deviated from known relationship and real law between data. Secondly, because the training data set is just a subset in sample space, if we can't discover its distribution, as well as sample size is not large enough, or enough data were collected, but they do not conform to its real distribution, even if there is no noise data, finally the prediction model can't accurately express real relationship and law between data. Thirdly, because the space where the final neural network is from is not known in advance, traditional deep learning algorithms can't guarantee the final neural network can express exactly the true relationship and the real law between data and difficultly ensure the model full compliance with professional knowledge. Finally, if the optimization problem is multi-peak complex function, recently there is no strong optimization mechanism to solve it effectively for the optimal interpretable prediction model. Thus, in order to improve the interpretability of the fully connected layers, the sample must be dense enough, its distribution must be accurately known, and interference of the noise data can be avoided easily, the learning network's space must be fully known, even prediction model posterior distribution $p(y_t|w, x)$ is clear and a good optimization algorithm is also essential. However, in fact, these strict conditions cannot be met.

### Logical Network

Fan et al. (2018) think that fuzzy logic can offer better interpretability in terms of logic inference rules for the complex neural network. In order to extract the logic relationship between input and output from the fully connected layers, this paper designs a general logic network. The general logic network structure is shown in the figure below.

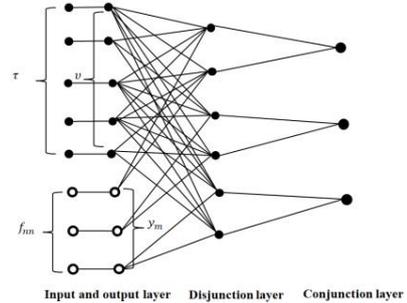

Fig.1 The general logic network structure

The general logic network consists of three layers of nodes. The first layer is the input and output layer, which is composed of feature predicate nodes (solid points in Fig.1) and decision result predicate nodes (hollow points in Fig.2). Each feature predicate is a predicate function about a feature map of the fully connected layers. Each decision result predicate is a predicate function of a result output of the fully connected layers. All feature predicate nodes and each decision result predicate node are connected to a disjunction node of the disjunction layer with true or false edges. Each disjunctive node represents a disjunctive paradigm obtained by a logical disjunction on some feature predicate nodes and a decision result predicate node. If a predicate node is connected to the disjunction node by a false edge, then the predicate corresponding to the predicate node is preceded by a non-operator in this disjunction paradigm. Otherwise there are no non-operators. In order to optimize the logical network structure, a membership degree of logic false is defined for each true or false edge. If the same decision result predicate node exists in multiple disjunctive paradigms, then these disjunctive paradigms should be joined to a conjunctive paradigm. Meanwhile, all of these disjunctive nodes corresponding to these disjunctive paradigms are connected to a conjunction node of the conjunction layer by true or false edges. Each conjunctive paradigm forms a group in the network, and each group has an eigenvalue in a world. The joint distribution of the world in the logical network can be obtained by using these eigenvalues.

The eigenvalue of each group in the logic network can be calculated using t-norm. The calculation method is shown in the following table.

Table 1 t-norm calculation method about an eigenvalue of each group

| t-norm opertation | Product | Minimum | Lukasiewicz |
|---|---|---|---|
| $a \wedge b$ | $a \cdot b$ | $min(a,b)$ | $max(0, a+b-1)$ |
| $a \vee b$ | $a+b-a \cdot b$ | $max(a,b)$ | $min(1, a+b)$ |
| $\neg a$ | $1-a$ | $1-a$ | $1-a$ |
| $a \Rightarrow b$ | $min\left(1, \frac{b}{a}\right)$ | $a \leq b ? 1 : b$ | $min(1, 1-a+b)$ |

Suppose the logic network contains a logical la $\forall \tau_1 \forall \tau_2 \neg A(\tau_1,) \vee \neg B(\tau_2) \vee C(y_{m1})$, where $A(\tau_1)$ and $B(\tau_2)$ are corresponding feature predicates of two input feature maps $\tau_1$ and $\tau_2$ of the fully connected layers respectively and $C(y_{m1})$ is a decision result predicate of an output result $y_{m1}$ of the fully connected layers. The eigenvalue of the formula can be obtained by using the Lukasiewicz method in Table 1.

$$\Phi_c(y_{m1}) = \frac{1}{|D_{\tau_1}||D_{\tau_2}|} min_{\tau_1 \in D_{\tau_1}, \tau_2 \in D_{\tau_2}}(1, a[1 - A(\tau_1)] + (1-a)A(\tau_1) + b[1 - B(\tau_2) + (1-b)B(\tau_2)] + c[1 - C(y_{m1})] + (1-c)C(y_{m1})) \quad (7)$$

where a, b and c are the membership degrees which true or false edges between predicate nodes of $A(\tau_1)$, $B(\tau_2)$ and $C(y_{m1})$ and disjunctive node of the disjunction layer is logic false. Their initial values $a = 1, b = 1, c = 0$. These nodes form a group. The other kind of group contains output nodes of fully connected layers and decision result predicate nodes of logical networks. Each group uses an edge to connect an output node of fully connected layers with a decision result predicate node. Its eigenvalue is

$$\Phi_d(y_m, f_{nn}) = |y_m - f_{nn}|^2 \quad (8)$$

When the weight vector of all eigenvalues is $\lambda$, the input and output of the fully connected layers are $\Gamma$ and $f_{nn}$ respectively, the conditional probability distribution function by which the closed logical network is true is

$$p(y_m|f_{nn}, \lambda, \Gamma) = \frac{1}{Z(y_m)} exp(-\Phi_d(y_m, f_{nn}) + \Sigma_i \lambda_i \Phi_c(y_{mi})) \quad (9)$$

By maximizing its likelihood function, the optimal value of decision result predicate $y_M$ and the optimal membership degrees $\gamma$ of true or false edges in the logical network can be obtained.

$$C(y_M) = \underset{y_m, \gamma}{argmax}[\log p(y_m|f_{nn}, \lambda, \Gamma)] \propto \underset{y_m, \gamma}{argmax}[-\Phi_d(y_m, f_{nn}) + \Sigma_i \lambda_i \Phi_c(y_{mi})] \quad (10)$$

An input of every feature predicate is a final feature map $\tau$ which is captured by the network in feature extraction stage. By calculating the similarity between the feature map $\tau$ and the template feature map $\upsilon$ of the feature predicate, the value of the predicate is obtained. If the similarity is equal to 0, the predicate value is 1; otherwise, the predicate value 0. Because $KL$ divergence has asymmetry, the similarity is obtained by $JS$ divergence between the feature map $\tau$ and the template feature map $\upsilon$.

$$JS(\tau_j \| \upsilon_i) = \frac{1}{2} KL\left(\upsilon_i \left\| \frac{\upsilon_i + \tau_j}{2}\right.\right) + \frac{1}{2} KL\left(\tau_j \left\| \frac{\upsilon + \tau_j}{2}\right.\right) \quad (11)$$

Its range is [0,1], where 0 is the same and 1 is different. The algorithm for solving the optimal decision result predicate $y_M$ and the optimal membership degree $\gamma$ of true or false edges of the logical network is as follows:

**Inputs**: feature map set $\Gamma$ obtained by the feature extraction stage, the template feature map set $V$ of the feature predicate, final output $f_{nn}$ of the fully connected layers, the weight vector $\lambda$ of all eigenvalues and the logic network $LN$.
**Outputs**: the optimal decision result predicate $y_M$ and the optimal membership degree vector $\gamma$ of true or false edges of $LN$.
**Initialization**: the membership $\gamma$ of True or False edge of $LN$.
**for** No converged or $i = 1$ **to** the maximum number $T$ of iterations **do**
  1) Compute the optimal $y_M$ with the logic network $LN$ by maximizing the likelihood function Eq. (10)
  2) Compute the optimal $\gamma$ based on $y_M$ by maximizing the likelihood function Eq. (10).
**end**

Algorithm 1: Iterative algorithm to solve the optimal decision result predicate $y_M$ and the optimal membership degree γ of true or false edges of the logical network.

## The game between perceptual network and logical network

In the process of extracting the logic relation between the input and the output of the fully connected layers from the perceptual network, the maximum probability $p(\theta|X, y_t)$ should be guaranteed, where the parameter vector $\theta$ includes $w$ and $\lambda$.

$$p(\theta|X, y_t) = \frac{p(\theta|X)p(y_t|\theta, X)}{p(y_t|X)} \propto p(\theta)p(y_t|\theta, X) \quad (12)$$

where
$p(y_t|\theta, X) =$
$\int p(y_t|f, \theta, X) p(f|w, X) \int p(f|y_m, \lambda, X) p(y_m|\lambda, X) dy_m \, df$

When the logical network is known, $p(y_M|\lambda, X) = 1$. Then

$$\int p(f|y_m, \lambda, X) p(y_m|\lambda, X) dy_m = p(f|y_M, \lambda, X) \quad (13)$$

Known the input $X$ and $w$, $p(f_{nn}|w, X) = 1$ where $f_{nn1}$ is the output of the convolutional neural network. Then

$$p(y_t|\theta, X) = \int p(y_t|f, \theta, X) p(f|w, X) p(f|y_M, \lambda, X) df = p(y_t|f_{nn}, \theta, X) p(f_{nn}|y_M, \lambda, X) \quad (14)$$

If the parameter vector $\theta$ of the convolutional neural network and training sample $X$ are given, the loss function between the target value $y_t$ of $X$ and the output $f_{nn}$ of the neural network obey the normal distribution. The distribution function is

$$p(y_t|f_{nn}, \theta, X) = \frac{exp(\Phi_r(y_t, f_{nn}))}{Z(y_t)} \quad (15)$$

Assuming that $X$, $\lambda$ and the membership degree $\gamma$ of the true or false edge of the logical network are known, the conditional probability distribution function of the output value $f_{nn}$ of the neural network on the logical network is

$$p(f_{nn}|y_M, \lambda, X) = \frac{exp(-\Phi_d(y_M, f_{nn}) + \sum_c \lambda_c \Phi_c(y_M))}{Z(f_{nn})} \quad (16)$$

where $Z(y_t)$ and $Z(f_{nn})$ usually can be constant. Then by maximizing a likelihood function of $p(\theta|X, y_t)$ the optimal network parameter vector $\theta$ can be obtained.

$C_\theta(X, y_t) = argmax_\theta [log \, p(\theta|X, y_t)]$
$= argmax_\theta [log \, p(w) + log \, p(\lambda) + \Phi_r(y_t, f_{nn}) - \Phi_d(y_M, f_{nn}) + \sum_c \lambda_c \Phi_c(y_{M1})] \quad (17)$

where $\theta$ include $w$ and $\lambda$, and

$$\Phi_r(y_t, f_{nn}) = -\frac{1}{2} \sum_l |y_t - f_{nn}|^2, \quad (18)$$

$$\Phi_d(y_M, f_{nn}) = \frac{1}{2} \sum_l |y_M - f_{nn} - \frac{1}{l} \sum_l |y_M - f_{nn}||^2. \quad (19)$$

## Game Algorithm of DCLM

The logical network should firstly be constructed based on the fully connected layers. When the first training sample is inputted into the convolution network, $N$ feature maps can be obtained. The $N$ feature maps are assigned to $N$ feature predicate nodes of a logical network as template feature maps and predicate values of the predicate nodes are 1. Each node is connected to $M$ disjunctive nodes by $M$ false edges, and $M$ decision result predicate nodes are connected to the $M$ disjunctive nodes by $M$ true edges respectively. Thus $M$ disjunctive paradigms can be obtained. Each disjunction node is connected to a junction node by a true edge, constructing $M$ conjunctive paradigms. After the second training sample is inputted into the convolution network, new $N$ feature maps can be gotten. By $JS$ divergence between the feature map and every template feature map, the feature map is inputted into the feature predicate node whose template feature map have the smallest $JS$ divergence which is less than the specified threshold than other nodes. Otherwise, we regenerate a feature predicate node for the feature map. Each node is connected to $M$ new disjunctive nodes by $M$ false edges, and $M$ decision result predicate nodes are connected to the $M$ new disjunctive nodes by $M$ true edges respectively. Thus we can obtain $M$ new disjunctive paradigms. Each new disjunctive node and an old disjunctive node which connect a same decision result predicate node with the new disjunctive node are connected to a same conjunctive node by true edges. Then $M$ new conjunctive paradigms can be obtained. And so on and so on, and eventually we get a complete logical network which save all logical relations between input and output of the fully connected layers.

The process of extracting a logical network from a fully connected network can be carried out by maximizing the likelihood of the training data. In particular, assuming that all parameter priors follow Gaussian distributions, we get:

$$C_\theta(X, y_t) = argmax_\theta \left[-\frac{\alpha}{2}\|w\|^2 - \frac{\beta}{2}\|\lambda\|^2 + \Phi_r(y_t, f_{nn}) - \Phi_d(y_M, f_{nn}) + \sum_c \lambda_c \Phi_c(y_M)\right] \quad (20)$$

where $\alpha_1$, $\alpha_2$ and $\beta$ are meta-parameters determined by the variance of the selected Gaussian distributions. Turn it into a minimization problem:

$$C_\theta(X, y_t) = argmin_\theta \left[\frac{\alpha}{2}\|w\|^2 + \frac{\beta}{2}\|\lambda\|^2 - \Phi_r(y_t, f_{nn}) + \Phi_d(y_M, f_{nn}) - \sum_c \lambda_c \Phi_c(y_M)\right] \quad (21)$$

Also in this case the likelihood may be maximized by gradient descent using the following derivatives on the model parameters $\lambda_c$ and $w$:

$$\frac{\partial C_\theta(X, y_t)}{\partial \lambda_c} = \beta \lambda_c - \Phi_c(y_M) \quad (22)$$

$$\frac{\partial C_\theta(X, y_t)}{\partial w} = \alpha_1 w_1 + \frac{\partial \sum_l |y_t - f_{nn}|}{\partial w} + \frac{\partial \Phi_d(y_M, f_{nn})}{\partial w} \quad (23)$$

The algorithm of iterative optimization algorithm is shown as follows:

**Inputs**: Input data $X$, output targets $y_t$.
**Outputs**: the optimal model parameters $w$ and the optimal logic network $LN$ and the optimal weight vector $\lambda_c$ of all eigenvalues of $LN$.
**Initialization**: all parameters $\lambda_c$ and $w$.
**for** No converged or $i = 1$ **to** the maximum number $T$ of iterations **do**
1) Compute function outputs $f_{nn}$, and feature maps $\Gamma$ using current function weights $w$.
2) Generate the logic network $LN$ with the feature maps $\Gamma$, using current weight vector $\lambda_c$ of all ei-

genvalues.

3) Comput the optimal $y_M$ with the logic network $LN$.
4) Compute gradient $\nabla_\theta C_\theta(X, y_t)$.
5) Update $\theta$ via gradient descent: $\theta_i = \theta_{i-1} - \mu \nabla_\theta C_\theta(X, y_t)$.

**end**

**Algorithm 2**: Game algorithm of DCLM.

## Experimental verification

We designed four experiments to evaluate the DCLM. The first experiment is to explore whether there is a logical relationship between the final feature maps obtained in the feature extraction stage of CNN and the network's optimal decision result. The second experiment is to carry out a test of a stability of extracting the logical relationships between feature maps and decision result of CNN using the game algorithm of DCLM. The third one is to verify the convergence performance of logical network optimization algorithm. The final one is to evaluate whether DCLM can improve the interpretable performance of the fully connected layer of CNN. In these experiments, the MNIST database of handwritten digits was used.

### Experiment 1: Verification of the existence of logical relations

In the experiment, 1000 random images were feed into complete CNN and uncomplete CNN which were deleted every one of 32 feature map inputs on its fully connected layer respectively.

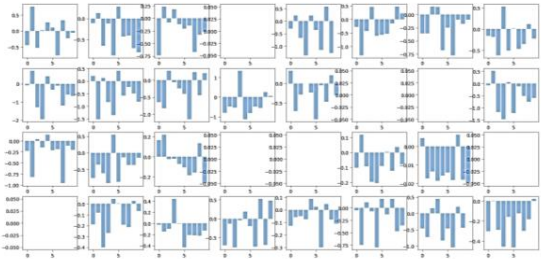

Fig.2 32 bar charts of the average difference between the two outputs on complete trained CNN and uncomplete trained CNN which were deleted every one of 32 feature map inputs on its fully connected layer respectively.

The average differences between the outputs of the two networks were calculated on each of the output dimensions. In trained CNN and untrained CNN, 32 bar charts were obtained respectively. In these bar charts the abscissa is the serial number of each dimension of the output of CNN, and the ordinate is the average difference value.

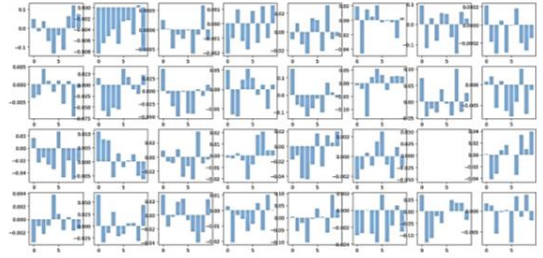

Fig.3 32 bar charts of the average difference between the two outputs on complete untrained CNN and uncomplete untrained CNN which were deleted every one of 32 feature map inputs on its fully connected layer respectively.

As seen from Fig.2 and Fig.3, most average difference values are not 0. Some values exceed 0.5. There are positive values and negative values. In a small number of charts we find all values are 0. This is mainly because these deleted feature maps is all empty. They have no impact on the network output. Meanwhile, we can also find that the direction and strength of the action of the same feature map on each dimension of the output of CNN are different. These phenomena on the trained CNN are more conspicuous than those on the untrained CNN. These results might prove that most non-empty feature maps are important to the outputs of trained CNN very much and the logic relationships between the feature maps and the outputs are complex, with positive correlation and negative correlation. It is very difficult to find the logical relationship between the feature maps and the outputs manually.

### Experiment 2: Stability test of logical relationship discovery

In this experiment, 1000 random graphs were used as training samples of CNN. During each training process, all input feature maps of the fully connected layer were extracted to form a feature map group. All the feature map groups were composed of a set. The number of feature map groups in the set was recorded in real time. The curve of the numbers of the feature map groups is shown in Fig. 4.

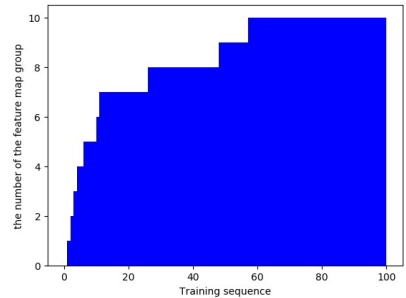

Fig.4 the curve of the numbers of the feature map groups with the increase of training times (the abscissa is the training times and the ordinate is the number of the feature map groups.)

The results show that with the increase of training times, the number of the feature map groups keeps increasing at the beginning, but when the number reaches a certain value, it tends to be stable. This indicates that the number of logical relationships between the feature maps and the outputs is limited.

## Experiment 3: Convergence test of logical network optimization algorithm

In a training process of DCLM, the objective function value after each optimization of the logical network *LN* was collected, and the result is shown in Fig. 5. The abscissa is the number of logical network optimization, and the ordinate is the objective function value.

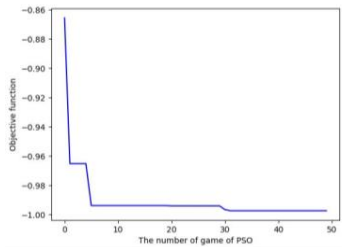

Fig.5 the curve of the objective function value with the increase of training times

As can be seen from Fig. 5, with the increase of optimization times, the objective function value continues to decrease, and then tends to be stable. The results show that the algorithm can converge in a finite time.

## Experiment 4: Performance evaluation of DCLM

In order to improve the experimental efficiency without affecting the conclusion, this experiment randomly selected 1000 images from MNIST data set, trained a traditional CNN containing 2 convolutional layers and a fully connected layer. The last convolutional layer output 32 feature maps. The DCLM was applied to the CNN, represented by DCLM-CNN. It is compared with three different CNN without DCLM(CNN1 has two convolution layers, CNN2 has three convolution layers, and CNN3 has four convolution layers). The main performance comparisons include: predictive accuracy, interpretable performance. The interpretable performance was evaluated from two aspects: one is to evaluate the average errors between the outputs of the fully connected layer and the outputs of the logical network under the same input; the other is to evaluate whether the algorithm can provide the logical relationship between the inputs and outputs of the fully connected layer.

Table 2 Experimental results for MNIST data sets

|  | Accuracy | Interpretation measure |
|---|---|---|
| CNNs-1 | 73.4 | NULL |
| DCLM-CNNs-1 | 66.55 | 0.47 |
| CNNs-2 | 78.43 | NULL |
| DCLM-CNNs-2 | 71.37 | 0.49 |
| CNNs-3 | 76.62 | NULL |
| DCLM-CNNs-3 | 74.33 | 0.05 |

As seen from Table 2, the accuracy value of DCLM-CNN is not significantly worse than the other algorithms. But the predication model by DCLM-CNN has very good interpretable performance. The main reason is that its predication results on the testing dataset are closed to the predication results of the logic network which was extracted from the fully connected layer of the CNN. This proves that the predication model obeys the logical relations in the logical network. Meanwhile, DCLM-CNN can provide the optimal membership degree matrix of true or false edges of logic network, which can describe well all logical relationships between the input feature maps and the result output of the fully connected layer of CNN. Partial results of the matrix are shown in Table 3.

Table 3 the optimal partial membership degree matrix(column headings are feature predicates and the row title is the disjunction node)

|  | A1 | A2 | A3 | A4 | A5 |
|---|---|---|---|---|---|
| Y1 | 0.645 | -1 | 1 | 1 | 0.645 |
| Y2 | -1 | 0.645 | -0.645 | -0.645 | -1 |
| Y3 | 0.645 | -0.992 | -1 | 0.645 | -1 |
| Y4 | -0.645 | -0.645 | -0.685 | -0.645 | -1 |
| Y5 | -1 | 0.645 | -0.645 | 0.606 | 0.645 |
| Y6 | -0.645 | -0.148 | 0.645 | -0.645 | 0.645 |
| Y7 | -0.092 | -1 | -0.645 | -0.355 | 0.645 |
| Y8 | 1 | 1 | -0.705 | -0.645 | -1 |
| Y9 | -0.645 | -1 | 1 | -0.645 | 1 |
| Y10 | 0.285 | 0.645 | -1 | -0.866 | -0.645 |

## Conclusion

In this paper, the interpretability of fully connected layers of CNN is focused on. This paper analyzes why the problem exists in fully connected layers of CNN. Then a method of extracting logical relation between input and output of fully connected layers, DCLM, is proposed. A game process between perceptual learning and logical abstract cognitive learning is implemented to improve the interpretable performance of deep learning process and deep learning model in the algorithm.

The following conclusions can be drawn from the experimental results. Firstly, the logical relationship between the feature map inputs and the outputs of fully connected layers of CNN is important to the prediction and so complex that and DCLM had significantly higher stability and we cannot find the logical relationship manually. Secondly, the number of logical relationships is limited. Thirdly, the algorithm which finds the logical relationships can converge in a finite time. Finally, the proposed algorithm not only demonstrates good prediction accuracy, but also extracts the logical relationship between input and output of the full connection layer, and finally provides the logical interpretation of the full connection layer of CNN.